\documentclass[10pt,twocolumn,letterpaper]{article}

\usepackage{cvpr}              

\usepackage{graphicx}

\usepackage{amsmath}
\usepackage{amssymb}
\usepackage{booktabs}
\usepackage{multirow}
\usepackage{xcolor,colortbl}
\usepackage{enumerate}
\usepackage{verbatim}
\usepackage{soul}
\usepackage{listings}
\usepackage{amssymb,algorithm}

\usepackage[pagebackref,breaklinks,colorlinks]{hyperref}

\usepackage[capitalize]{cleveref}
\usepackage[accsupp]{axessibility}
\crefname{section}{Sec.}{Secs.}
\Crefname{section}{Section}{Sections}
\Crefname{table}{Table}{Tables}
\crefname{table}{Tab.}{Tabs.}

\def\cvprPaperID{5399} 
\def\confName{CVPR}
\def\confYear{2022}

\newcommand{\wref}{\bar{w}}

\newcommand{\bs}{\mathbf{s}}
\newcommand{\bw}{\mathbf{w}}

\newcommand{\PB}{Piggyback}
\newcommand{\RA}{Res. Adapt.}
\newcommand{\DA}{Deep Adapt.}

\newcommand{\DWS}{TAPS}
\newcommand{\DWSF}{Task Adaptive Parameter Sharing}

\newcommand{\Flowers}{Flowers}
\newcommand{\Cars}{Cars}
\newcommand{\Wiki}{WikiArt}
\newcommand{\Sketch}{Sketch}
\newcommand{\CUB}{CUB}
\newcommand{\DN}{DenseNet-121}
\newcommand{\RN}[1]{ResNet-#1}

\renewcommand{\paragraph}[1]{\vspace{.5em}\noindent\textbf{#1.}}

\begin{document}

\title{\DWSF{} for Multi-Task Learning}


\renewcommand{\thefootnote}{\fnsymbol{footnote}} 
\author{
Matthew Wallingford$^1$ \thanks{Work done during an internship at AWS AI Labs.}
\quad Hao Li$^2$
\quad Alessandro Achille$^2$\\
\quad Avinash Ravichandran$^2$ \thanks{Corresponding Author.} 
\quad Charless Fowlkes$^2$
\quad Rahul Bhotika$^2$
\quad Stefano Soatto$^2$ \\
\texttt{\footnotesize mcw244@cs.washington.edu} \quad
\texttt{\footnotesize \{haolimax,achille,ravinash,fowlkec,bhotikar,soattos\}@amazon.com} \\
{\small 
$^{1}$University of Washington
\quad $^{2}$AWS AI Labs
}
}
\maketitle

\begin{abstract}
Adapting pre-trained models with broad capabilities has become standard practice for learning a wide range of downstream tasks. The typical approach of fine-tuning different models for each task is performant, but incurs a substantial memory cost. To efficiently learn multiple downstream tasks we introduce \DWSF{}~(\DWS{}), a general method for tuning a base model to a new task by adaptively modifying a small, task-specific subset of layers. This enables multi-task learning while minimizing resources used and competition between tasks. \DWS{} solves a joint optimization problem which determines which layers to share with the base model and the value of the task-specific weights.
Further, a sparsity penalty on the number of active layers encourages weight sharing with the base model. Compared to other methods, \DWS{} retains high accuracy on downstream tasks while introducing few task-specific parameters. Moreover, \DWS{} is agnostic to the model architecture and requires only minor changes to the training scheme. We evaluate our method on a suite of fine-tuning tasks and architectures (ResNet, DenseNet, ViT) and show that it achieves state-of-the-art performance while being simple to implement.
\end{abstract}
\begin{figure*}[ht!]
\centering
\begin{minipage}[c]{.21\linewidth}
\centering
    \includegraphics[width=.99\linewidth]{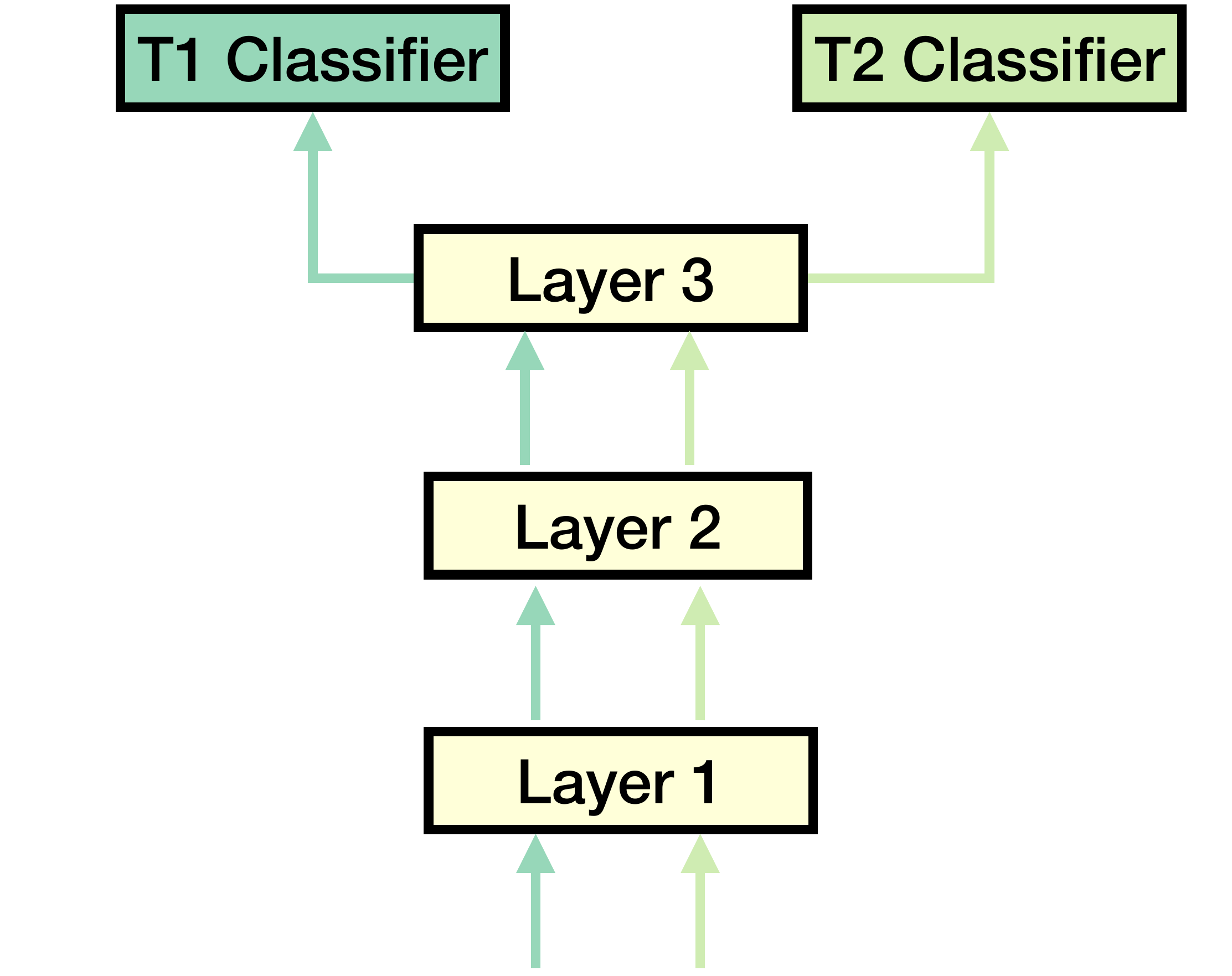}
(a) Feature Extractor
\end{minipage}
\begin{minipage}[c]{.21\linewidth}
\centering
    \includegraphics[width=.99\linewidth]{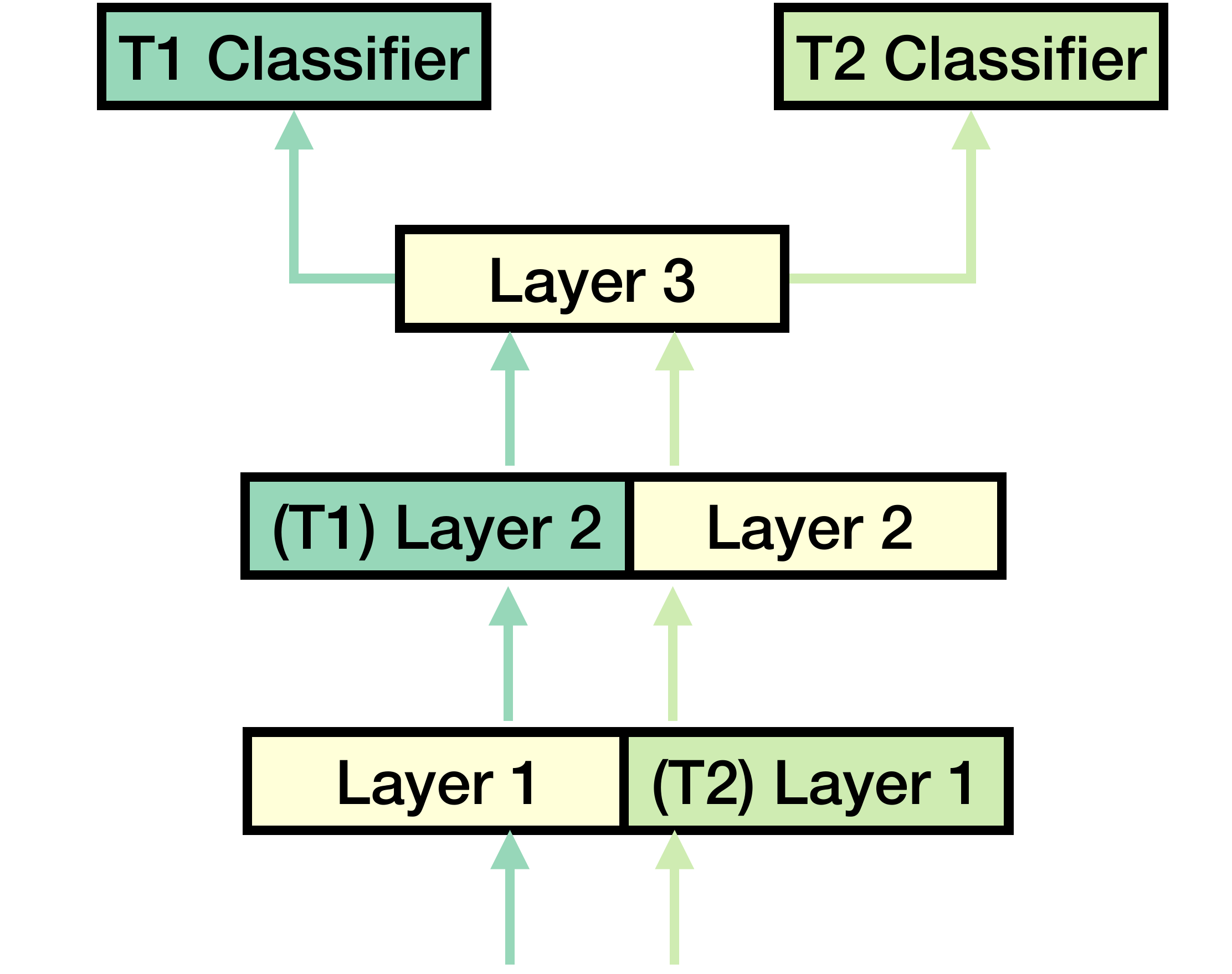}
   (b) Our Method
\end{minipage}
\begin{minipage}[c]{.21\linewidth}
\centering
    \includegraphics[width=.99\linewidth]{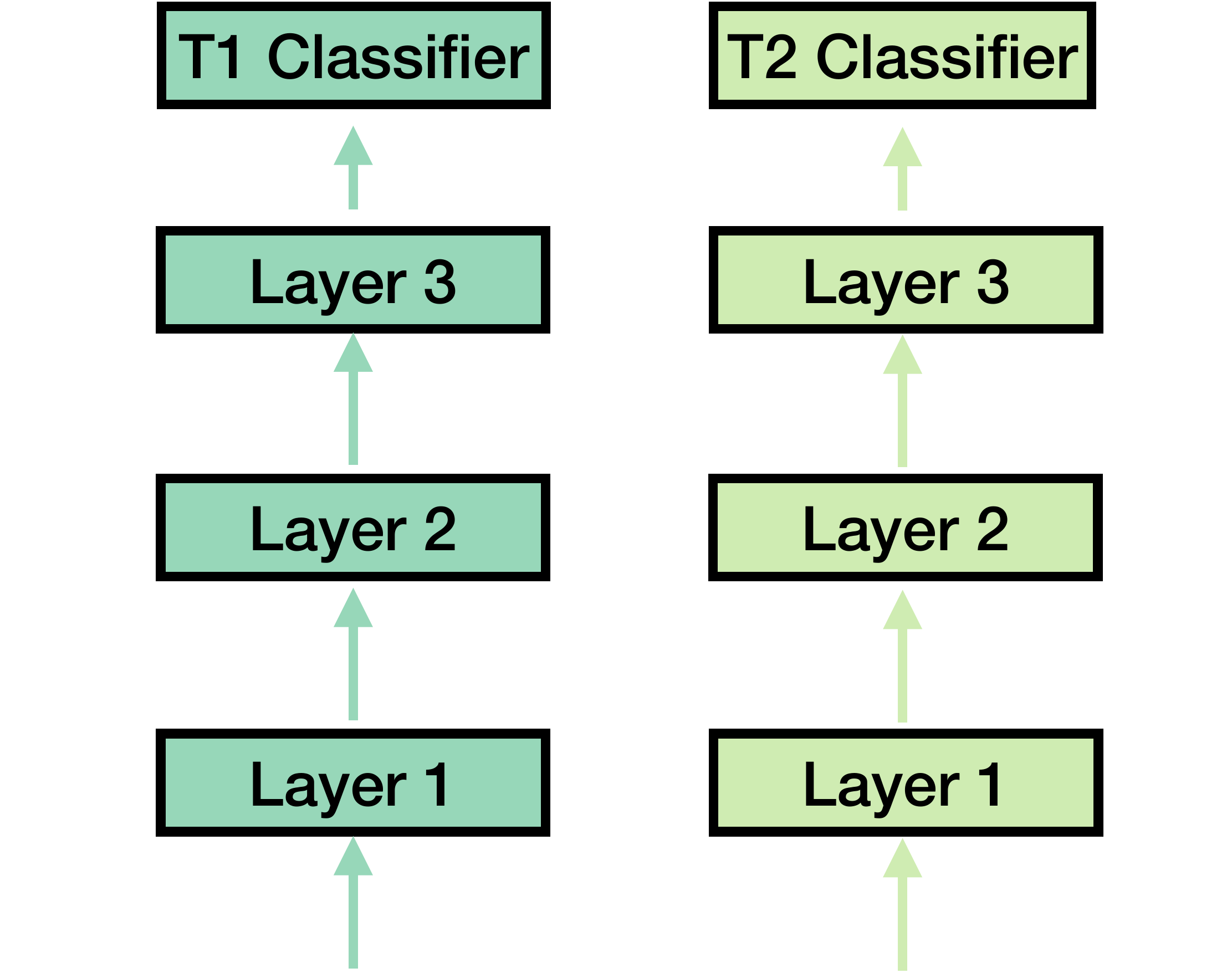}
      (c) Fine-Tuning
\end{minipage}

    \caption{{\textbf{Overview of our approach.} Difference between (b) our approach, (a) feature extractor as well as (c) finetuning. Here we have two tasks $T_1$ and $T_2$. $T_1$ shown by turquoise and $T_2$ by green. Yellow boxes denote the base network layer. Our approach adds task specific parameters to different layers based on the target tasks. Notice that this is in contrast to (a) no task specific layers are used and (c) where every layer is tasks specific. Also (c) suffers from catastrophic forgetting, and lose the base network's parameters unlike (b) and (a).}}

    \label{fig:teaser}
\end{figure*}
\section{Introduction}

Real-world applications of deep learning frequently require performing multiple tasks (Multi-Task Learning/MTL). To avoid competition between tasks, a simple solution is to train separate models starting from a common pre-trained model. Although this approach results in capable task-specific models, the training, inference, and memory cost associated grows quickly with the number of tasks. Further, tasks are learned independently, missing the opportunity to share when tasks are related.

Ideally, one would train a single model to solve all tasks simultaneously. A common approach is to fix a base model and add task-specific parameters (e.g., adding branches, classifiers) which are trained separately for each task. However, deciding where to branch or add parameters is non-trivial since the optimal choice depends on both the initial model and the downstream task, to the point that some methods train a secondary network to make these decisions. 

Moreover, adding weights (layers, parameters, etc.) to a network independent of the task is also not ideal: some methods ~\cite{mallya2018piggyback, rebuffi2017learning, li2016revisiting} add a small, fixed number of learnable task-specific parameters, however they sacrifice performance when the downstream task is dissimilar from the pre-training task. Other methods perform well on more complex tasks, but add an excessive number of task-specific parameters even when the downstream task is simple
\cite{Rusu2016ProgressiveNN, guo2019spottune, zhang2020side} 

In this work, we overcome these issues by introducing \DWSF{}~(\DWS{}).
Rather than modifying the architecture of the network or adding a fixed set of parameters, \DWS{} adaptively selects a minimal subset of the existing layers and retrains them.
At first sight, selecting the best subset of layers to adapt is a complex combinatorial problem which requires an extensive search among $2^L$ different configurations, where $L$ is the number of layers. 
The key idea of \DWS{} is to relax the layer selection to a continuous problem, so that deciding which layers of the base model to specialize into task-specific layers can be done \textit{during training} by solving a joint optimization using stochastic gradient descent.

The final result is a smaller subset of task-specific parameters (the selected layers) which replace the base layers. Our approach has several advantages: (i) It can be applied to any architecture and does not need to modify it by introducing task-specific branches; (ii) \DWS{} does not reduce the accuracy of the target task (compared to the paragon of full fine-tuning) while introducing fewer task-specific parameters; (iii) The decision of which layers to specialize is interpretable, done with a simple optimization procedure, and does not require learning a policy network; (iv) It can be implemented in a few lines, and requires minimal change to the training scheme.

Our method finds both intuitive sharing strategies, and other less intuitive but effective ones.  For example, \DWS{} tends to modify the last few layers for ResNet models while for Vision Transformers \DWS{} discovers a significantly different sharing pattern, learning to adapt only the self-attention layers while sharing the feed-forward layers.

We test our method on standard benchmarks and show that it outperforms several alternative methods. Moreover, we show that the results of our method are in-line with the standard fine-tuning practices used in the community. 
The contributions of the paper can be summarized as follows:
\begin{enumerate}

\item We propose \DWS{}, a method for differentiably learning which layers to tune when adapting a pre-trained network to a downstream task. This could range from adapting or specializing an entire model, to only changing $0.1\%$ of the pre-trained model, depending on the complexity/similarity of the new task. 
 

\item We show that \DWS{} can optimize for accuracy or efficiency and performs on par with other methods. Moreover, it  automatically discovers effective architecture-specific sharing patterns as opposed to hand-crafted weight or layer sharing schemes.

\item \DWS{} enables efficient incremental and joint multi-task learning without competition or forgetting. 
\end{enumerate}

\section{Related Work} 

\paragraph{Multi-Domain and Incremental Multi-Task Learning} 
In many applications, it is desirable to adapt one network to multiple visual classification tasks or domains (Multi-Domain Learning, or MDL). Unlike Multi-Task Learning (MTL) where the tasks are learned simultaneously, in MDL the focus is to learn the domains incrementally, as often not all data is available at once. Accordingly, in this work, we also refer to MDL as incremental MTL. The standard approach for adapting a network to a single downstream task is fine-tuning. However, adapting to multiple domains incrementally poses the challenge of catastrophically forgetting previously learned tasks. To foster research in the area, Rebuffi et al.~\cite{rebuffi2017learning} introduced the Visual Decathlon challenge and proposed residual adapters. Residual adapters fix most of the network while training small residual modules that adapt to new domains. This architecture was modified to a parallel adapter architecture in \cite{rebuffi2018efficient}. A controller based method called Deep Adaptation (DA) was introduced in \cite{rosenfeld2018incremental} to modify the learning algorithm using existing parameters. A simpler approach of using binary masks was proposed in Piggyback \cite{mallya2018piggyback}. Task specific masks are learned then applied to the weights of the original network. This approach was further extended in Weight Transformations using Binary Masks (WTPB) \cite{mancini2018adding} by modifying how the masks are applied. These methods focus on adding a small number of new parameters per task and underperform on more complex tasks as they have fixed capacity. Other solutions such as SpotTune~\cite{guo2019spottune} focus on performance without consideration for parameter efficiency. It trains an auxiliary policy network that decides whether to route each sample through a shared layer or task-specific layer. In contrast, \DWS{} does not require modification of the network architecture via adapters or training an auxiliary policy network. \DWS{} trains in one run with the same base architecture and requires no additional inference compute.

\paragraph{Parameter Efficient Multi-Domain Learning (MDL)}
Another line of work in MDL is that of parameter sharing~\cite{morgado2019nettailor,mallya2018packnet}. These approaches typically perform multi-stage training. NetTailor~\cite{morgado2019nettailor} leverages the intuition that simple tasks require smaller networks than more complex tasks. They train teacher and student networks using knowledge distillation and a three-stage training scheme.
PackNet~\cite{mallya2018packnet} adds multiple tasks to a single network by iterative pruning. However, pruning weights generally causes performance degradation. More recently,  Berriel \etal proposed Budget-Aware Adapters (BA$^2$) \cite{Berriel_2019_ICCV}. This method selects and uses feature channels that are relevant for a task. Using a budget constraint, a network with the desired complexity can be obtained. In summary,  most efficient parameter methods obtain efficiency at the loss of performance. Even with the largest budget in  $BA^2$, the performance is substantially lower compared to \DWS{}. Unlike existing methods, \DWS{} does not need to choose a high accuracy or a high efficiency regime. As shown in \cref{fig:pareto}, for a given task we can obtain models anywhere along the accuracy-efficiency frontier. 

\paragraph{Multi-Task Learning (MTL)}
 MTL focuses on learning a diverse set of tasks in the same visual domain simultaneously by sharing information and
computation, usually in the form of  layers shared across all tasks and specialized branches for specific tasks~\cite{kokkinos2017ubernet, ranjan2017hyperface}. A few methods have attempted to learn multi-branch network architectures~\cite{lu2017fully, vandenhende2019branched} and some methods have sought to find sharing parameters among task-specific models~\cite{misra2016cross, ruder2019latent,gao2019nddr}.
A closely related work is AdaShare~\cite{sun2019adashare}, which learns  a task-specific policy that selectively chooses which layers to execute for a given task in the multitask network. 
They use Gumbel Softmax Sampling \cite{maddison2016concrete,jang2016categorical} to learn the layer sharing policy jointly with the network parameters through standard back-propagation.
Since this approach skips layers based on the task, it can only be applied to a subset of existing architectures where the input and output dimension of each layer is constant. Additionally because the sharing policy is trained jointly across all tasks, AdaShare cannot learn in the incremental setting.

\paragraph{Incremental Learning}
Related to MDL is the problem of incremental learning. Here, the goal is to start with a few classes and incrementally learn more classes as more data becomes available. There are two approaches in this regard, methods that add extra capacity \cite{Rusu2016ProgressiveNN}, \cite{zhang2020side} (layer, filter, etc.) and methods that do not\cite{AGEM,EWC,icarl, li2017learning}. Methods that do not add extra capacity mitigate catastrophic forgetting by either using a replay buffer \cite{AGEM,icarl} or minimizing changes to the weights \cite{EWC}. Similar to our method, \cite{Rusu2016ProgressiveNN,zhang2020side} add capacity to the network to accommodate new tasks and prevent catastrophic forgetting. Progressive network \cite{Rusu2016ProgressiveNN} adds an entire network of parameters while Side-Tune \cite{zhang2020side} adds a smaller fixed-size network. Adding fixed capacity independent of the downstream task is sub-optimal since less complex tasks require fewer added parameters. In the extreme case where the pretrain task matches the downstream task no additional parameters are needed. In contrast to previous methods, \DWS{} adaptively adds capacity based on the downstream task and base network. 
Further, the goal of \DWS{} differs from incremental learning in that it starts with a pre-trained base model and learns new tasks or domains as opposed to adding new classes from a similar domain.

\section{Approach}

Given a pre-trained deep neural network with $L$ layers of weights and a set of $K$ target tasks $\mathcal{T}=\{T_1, T_2, ..., T_K\}$, for each task we want to select the minimal necessary subset of layers that needs to be tuned to achieve the best (or close to the best) performance. This allows us to learn new tasks incrementally while adding the fewest new parameters. 
In principle, this requires a combinatorial search over $2^L$ possible subsets. The idea of \DWSF{} (\DWS{}) is to relax the combinatorial problem into a continuous one, which will ultimately give us a simple joint loss function to find both the optimal task-specific layers to tune and optimize parameters of those layers. An overview of our approach is shown in \cref{fig:teaser}. 

\paragraph{Weight parametrization} We first introduce a  scoring parameter $s_i$ for each shared layer, where $i=1,\ldots ,L$. We then reparametrize the weights of each layer as:

\begin{equation}
    w_i = \wref_i + I_\tau(s_i) \, \delta w_i,
    \label{eq:dws}
\end{equation}
where $\wref_i$ are the (shared) weights of the pre-trained network and $\delta w_i$ is a trainable parameter which describes a task-specific perturbation of the base network. The crucial component is the indicator function $I_\tau(s_i)$ defined by
\begin{equation} 
I_\tau(s_i) = \left\{ 
  \begin{array}{ c l }
    1 & \quad \textrm{if } s_i \geq \tau, \\
    0                 & \quad \textrm{otherwise.}
  \end{array}
\right.
\end{equation}
for some threshold $\tau$. Hence, when $s_i \geq \tau$, the layer is transformed to a task-specific layer, and consequently will make its parameters task-specific. On the other hand, for $s_i < \tau$ the layer is the same as the base network and no new parameters are introduced.

The same approach can be used for different architecture and layer types, whether linear, convolutional, or attention. In the latter case, we add task-specific parameters to the query-key-value matrices as well as the projection layers.

\paragraph{Joint optimization} Given the parametization in \cref{eq:dws}, we can recast the initial combinatorial problem as a joint optimization problem over weight deltas and scores:
\begin{equation}
\label{eq:joint-optimization}
    (\delta \bw, \bs) = \arg\min_{\bw, \bs} \mathcal{L}_D(\bw) + \frac{\lambda}{L} \sum_{i=1}^{L}|s_i|,
\end{equation}
where $\bs=(s_1, \ldots, s_L)$, $\delta \bw=(\delta w_1,\ldots,\delta w_L)$, $\bw = (w_1, \ldots, w_L)$ and we denote with $L_D(\bw)$ the loss of the model on the dataset $D$. The first term of \cref{eq:joint-optimization} tries to optimize the task specific parameters to achieve the best performance on the task, while the second term is a sparsity inducing regularizer on $\bs$ which penalizes large values of $s_i$, encouraging layer sharing rather than introducing task-specific parameters.

\paragraph{Straight-through gradient estimation}
While the optimization problem in \cref{eq:joint-optimization} captures the original problem, it cannot be directly optimized with stochastic gradient descent since the gradient of the indicator function $I_\tau(s)$ is zero at all points except $\tau$. To make the problem learnable, we utilize a straight-through gradient estimator \cite{bengio2013estimating}. That is, we modify the backward pass and redefine the gradient update as:
\[
\nabla_{s_i} w_i = \delta w_i,
\]
in place of the true gradient $\nabla_{s_i} w_i = 0$. This corresponds to computing the derivative of the function $w_i = \wref_i + s_i \delta w_i$ rather than  $w_i = \wref_i + I_\tau(s_i) \delta w_i$. The inspiration to utilize the straight-through estimator for gating task-specific weight perturbations, $\delta w_i$, comes from sparsity literature \cite{wortsman2019discovering, rastegari2016xnor} where it has been used to prune network weights.

\paragraph{Joint MTL}
A natural question is if, rather than using a generic pretrained model, we can learn a base network optimized for multi-task learning. In particular, is there a pretrained representation that reduces the number of task-specific layers that need to be learned to obtain optimal performance?
To answer the question, we note that if data from multiple tasks is available simultaneously at training time, we can optimize \cref{eq:dws} jointly across all downstream tasks for both the base weights $\wref_i$ (which will be shared between all tasks) and the task specific $\delta w_i$. The loss function becomes:
\begin{align}
\label{eq:multitask-optimization}
    &(\mathbf{\bar{w}}, \delta \bw_1, \ldots , \delta \bw_K, \bs_1, \ldots, \bs_K) =\nonumber\\ &\quad\quad\arg\min_{\mathbf{\bar{w}},\delta \bw^k, \bs} \sum_{k=1}^{K} \Big( \mathcal{L}_{D_k}(\bar{\bw}, \delta \bw^k) + \frac{\lambda}{L}   \sum_{i=1}^{L}|s_i^{k}| \Big),
\end{align}
where $K$ is the total number of tasks, $\bar{\bw}$ is shared between all tasks and $\delta \bw_k$ and $\bs_k$ are task specific parameters. This loss encourages learning common weights $\bar{\bw}$ in such a way that the number of task specific parameters is minimized, due to the $L_1$ penalty on $\bs$. In Sec.~\ref{sec:mtl} we show that the joint multi-task variant of \DWS{} does increase weight sharing without loss in accuracy. In particular, the number of task-specific parameters is significantly reduced in the joint multi-task training setting with respect to incremental multi-task training.

A limitation of the joint multi-task variant of \DWS{} and other joint MTL methods \cite{sun2019adashare} is that the memory footprint during training increases linearly with the number of tasks. Our solution is to learn a single network which is trained jointly on all tasks, with task specific classifiers. Then train with the incremental variant of \DWS{} (Eq. \ref{eq:joint-optimization}) to adapt the jointly trained base network to each task. This approach achieves comparable accuracy and parameter sharing with the joint variant, while requiring constant memory during training. For comparison between the standard formulation and memory efficient variation see appendix \ref{sec:JointVariant}.

\paragraph{Batch Normalization}
Learning task specific batch normalization layers improves accuracy on average by $2-3\%$, while adding only a small amount of parameters ($.06\%$ for \RN{34} model). For this reason,  we follow the same setup as most methods and learn task-specific batch-norm parameters.


\section{Experiments}
In this section, we compare \DWS{} with existing methods in two settings: incremental MTL (\cref{sec:imtl}) and joint MTL (\cref{sec:mtl}). The details are as follows.  



\subsection{Incremental MTL \label{sec:imtl}}
In this scenario, methods adapt the pre-trained model for each task individually and combine them into a single model that works for every domain.  This approach is efficient during training in terms of both speed and memory as it can be parallelized and requires at most 2$\times$ the parameters of the base model. Alternatively, all tasks could interact and learn common weights, which is the joint multi-task scenario described in \cref{sec:mtl}.

\paragraph{Datasets} 
We show results on two benchmarks. One is the standard benchmark used in \cite{guo2019spottune,mallya2018piggyback,mallya2018packnet,mancini2018adding} which consists of 5 datasets: \Flowers{} \cite{flowers}, \Cars{} \cite{cars}, \Sketch{} \cite{Eitz2012HowDH}, \CUB{} \cite{birds} and \Wiki{} \cite{Saleh2015LargescaleCO}. Following \cite{Berriel_2019_ICCV}, we refer to this benchmark as ImageNet-to-Sketch. For dataset splits, augmentation, crops, and other aspects, we use the same setting as \cite{mallya2018piggyback}. 
Our second benchmark is the Visual Decathlon Challenge~\cite{rebuffi2017learning}. This challenge consists of 10 tasks which include the following datasets:  ImageNet~\cite{Russakovsky2015ImageNetLS}, Aircraft~\cite{aircrafts},  CIFAR-100~\cite{krizhevsky2009learning},  Describable textures~\cite{dtd}, Daimler pedestrian classification~\cite{munder2006experimental},  German traffic signs~\cite{stallkamp2012man},  UCF-101 Dynamic Images~\cite{soomro2012ucf101},  SVHN~\cite{svhn},  Omniglot~\cite{lake2015human}, and Oxford Flowers~\cite{flowers}. For details about the datasets and their augmentation see appendix \ref{app:datasets}.

\paragraph{Methods of Comparison} 
Our paragon is fine-tuning the entire network separately for each task, resulting in the best performance at the cost of no weight sharing. Our baseline is the fixed feature extractor, which typically gives the worst performance and shares all layers. In the incremental multi-task setting we compare our method with Piggyback~\cite{mallya2018piggyback}, SpotTune~\cite{guo2019spottune}, PackNet~\cite{mallya2018packnet}, and Residual Adapters~\cite{rebuffi2018efficient}.

\paragraph{Metrics}
We report the top-1 accuracy on each task and the S-score for the Visual Decathlon challenge as proposed in \cite{rebuffi2017learning}. In addition, we report the total percentage of additional parameters and task specific layers needed for all tasks. The individual parameter counts are available in \ref{app:VisualDec}.
Methods \cite{mallya2018piggyback,mallya2018packnet, mancini2018adding} that use a binary mask for their algorithm report the theoretical total number of bits (e.g., 32 for floats, 1 for boolean) required for storage rather than reporting the total number of parameters. However, as \cite{mallya2018piggyback} notes, depending on the hardware, the actual storage cost in memory may vary (e.g., booleans are usually encoded as 8-bits).
To establish  parity between different reporting structures, we report the total number of parameters used without normalization and the normalized count (assuming that a boolean parameter can be stored as 8-bits). 


\begin{figure}[h]
    \centering
    \includegraphics[width=.97\linewidth]{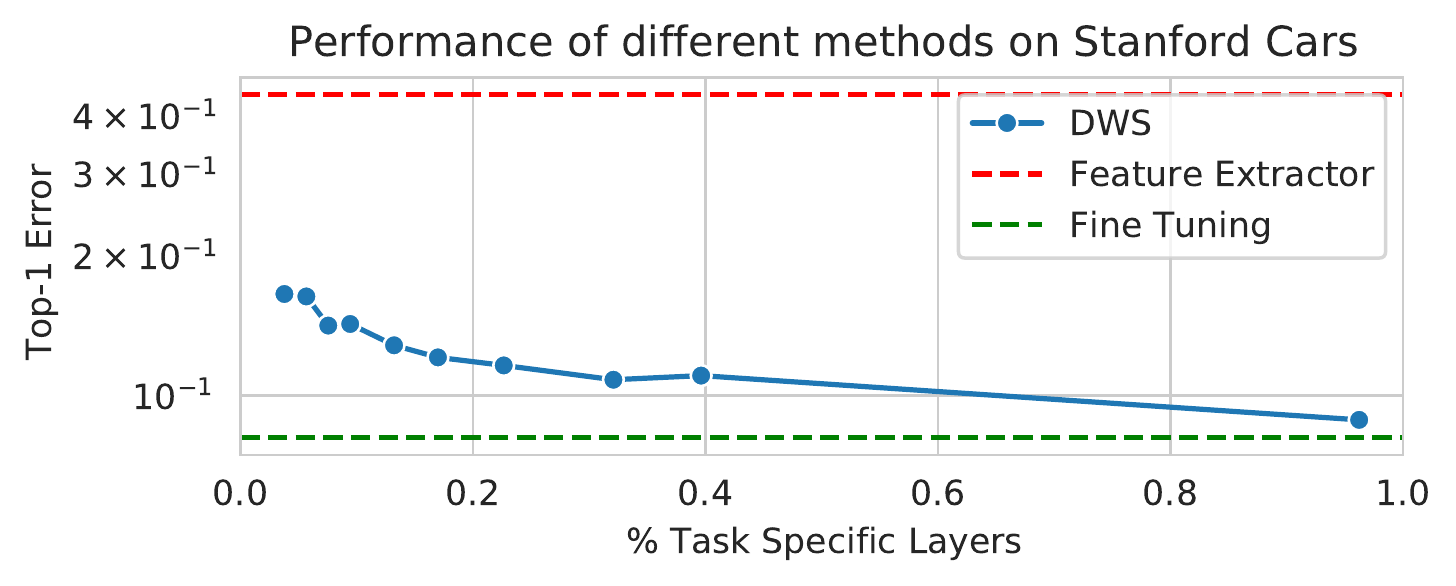}
    \caption{\textbf{Accuracy vs Task Specific Layers:} Accuracy vs percentage of task specific layers for the  \Cars{} dataset is shown. Varying $\lambda$ gives a variety of configuration. With even a very few task specific layers (high $\lambda$), we perform significantly better than the feature extraction baseline. Our method  also reaches the fine-tune performance but needs a significant number of task specific layers.}
    \label{fig:pareto}
\end{figure}

\paragraph{Training details} 
We use  an ImageNet pre-trained ResNet-50 \cite{He2016DeepRL} model as our base model for ImageNet-to-Sketch. We train \DWS{} for 30 epochs with batch size 32 on a single GPU. For the fine-tuning paragon and fixed feature extractor we report the best performance for the learning rates  $lr \in \{0.001, 0.005\}$. 

For \DWS{} use the SGD optimizer with no weight decay. We sweep over $\lambda \in \{0.25,0.5,0.75\}$ and $lr \in \{0.001, 0.005\}$. We fix the  threshold, $\tau=0.1$, for all datasets and use the cosine annealing learning rate scheduler. 

In addition to \RN{50}, we also apply \DWS{} to \DN{}~\cite{densnet} and Vision Transformers~\cite{dosovitskiy2020image}. To the best of our knowledge, we are the first to provide results for a transformer based architecture. The details of the training for these settings can be found in the \ref{app:incMTL}.


For the Visual Decathlon challenge, we use the WideResNet-28 as in \cite{rebuffi2017learning}, which is also referred to as \RN{26} in \cite{guo2019spottune}. Following existing work, we use a learning rate of 0.1 with weight decay of 0.0005 and train the network for 120 epochs. We report our results for  $\lambda \in \{0.25,0.5,1.0\}$. Like existing methods \cite{guo2019spottune, rebuffi2017learning, rebuffi2018efficient}, we report our accuracy on the test set while training on the training and validation dataset. We also calculate the $S$-Score to make a consolidated ranking of our method.

\paragraph{Results on ImageNet-to-Sketch}
\cref{tab:base} shows the comparison of our method with existing approaches on ImageNet-to-Sketch. Average accuracy over 3 runs of our method and fine-tuning are reported.  \DWS{} outperforms \PB{} and Packnet across all 5 datasets, Spot-Tune for 3 out of the 5 datasets, WTPB and BA$^2$ for 4 out of 5 datasets. We also note that \DWS{} uses, on average, only 57\% of the parameters that Spot-Tune does. We do not outperform existing methods on the \Cars{} dataset. Indeed, for this dataset the best results are obtained with $\lambda=0.0$ (see \cref{fig:pareto}), suggesting that most layers needs to be adapted for optimal performance. We report the number of parameters used for each task in appendix \ref{app:VisualDec}. In general, we perform significantly better in terms of accuracy compared to methods that are parameter efficient, while we achieve the same performance as methods that are designed for accuracy, but at a fraction of the parameter cost.


\begin{table*}[h]
\centering
\footnotesize
\caption{ \textbf{Performance of various method using a \RN{50} model on ImageNet-to-Sketch benchmark}. Accuracy for different methods are shown for the  different datasets. For \DWS{} and fine-tuning, we report the average accuracy over three runs. We report the total number of paramters (when available) and  in  parenthesis the data-type normalized parameter count. Numbers in bold denote the best performing method (other than fine-tuning) for each dataset. For Packnet, the arrow indicates the order of adding tasks. 
}
\setlength{\tabcolsep}{4pt}
\begin{tabular}{l|l|ccccc}
\toprule
 & Param Count & \Flowers & \Wiki & \Sketch  & \Cars  & \CUB\\
\midrule
Fine-Tuning & 6$\times$ & 95.73 & 78.02  & 81.83 & 91.89	 & 83.61	 \\
Feature Extractor & 1$\times$ & 89.14 & 61.74	 & 65.90 & 55.52 & 63.46 \\
\midrule
Piggyback~\cite{mallya2018piggyback}  & 6$\times$~(2.25$\times$) &  94.76&71.33&79.91&89.62&81.59  \\ 
Packnet \cite{mallya2018packnet} $\rightarrow$& (1.60$\times$)  &93.00&69.40&76.20&86.10&80.40\\
Packnet \cite{mallya2018packnet} $\leftarrow$ & (1.60$\times$)  & 90.60&70.3&78.7&80.0&71.4 \\
Spot-tune~\cite{guo2019spottune} &7$\times$~(7$\times$) & 96.34&75.77&80.2&\textbf{92.4}&\textbf{84.03} \\
WTPB \cite{mancini2018adding} & 6$\times$~(2.25$\times$) & 96.50&74.8&80.2&91.5&82.6 \\
BA$^{2}$ \cite {Berriel_2019_ICCV} & 3.8$\times$~(1.71$\times$) &95.74&72.32&79.28&92.14&81.19\\
\midrule 
\DWS{} & 4.12$\times$ & \textbf{96.68}& \textbf{76.94}&\textbf{80.74}&89.76&82.65
\\
\bottomrule
\end{tabular}
\label{tab:base}
\end{table*}

\begin{table}
\small
\caption{\textbf{Percentage of additional parameters and layers.} The percentage of task specific parameters and layers needed for each dataset across network architectures is shown. Numbers in bold denote the lowest across architectures. ViT-S/16 uses the least number of extra parameters, while \RN{50} adds the least number of task specific layers. }
\begin{tabular}{llllll}
\toprule
                       & \Flowers & \Wiki & \Sketch & \Cars & \CUB \\ \hline
\multicolumn{4}{l}{\textbf{Percentage of Additional Parameters}} & \\
\DN & 80.2&41.2&58.5&50.4&43.8\\
ViT-S/16 & \textbf{41.3}&\textbf{30.4}&\textbf{24.1}&\textbf{26.1}&\textbf{41.3}     \\ 
\RN{50}  &  65.5&52.8&75.9&41.9&70.6\\
\hline
\multicolumn{4}{l}{ \textbf{Percentage of Task Specific Layers}} &  \\
\DN &  69.4&22.5&41.1&28.3&23.9\\
ViT-S/16 & 54.2&20.8&22.9&37.5&54.2     \\ 
\RN{50}   & \textbf{22.6}&\textbf{20.8}& \textbf{43.4}&\textbf{14.5}&\textbf{28.3} \\
\hline
\end{tabular}
\label{tab:arch-param}
\end{table}

\begin{figure*}[h!]
    \centering
    \includegraphics[width=.93\linewidth]{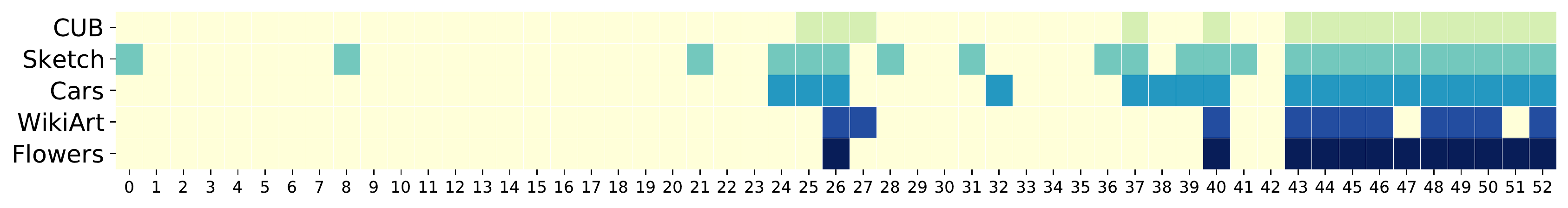}
    \caption{{\textbf{Task-specific layers for different datasets.} Each row shows the 53 convolution layers of \RN{50} layers for different datasets. Layer 0 is closest to the input, while layer 52 is closest to the classifier. Layers shown in yellow are shared with the ImageNet pre-trained model while layers shown in color are task-specific weights. We see that most tasks specific layers are toward the classifier.}}
    \label{fig:sharing-res}
\end{figure*}

\paragraph{Task-specific layers} In \cref{fig:sharing-res}, we show the layers that are task-specific for the different datasets. As expected, the final convolutional layers are always adapted. This corresponds to the common practice of freezing the initial 3 out of the 4 blocks of the \RN{50} model and fine-tuning the last block. But, interestingly, we see that some of the middle layers are also always active. For instance, layer 26 is often adapted as a task-specific layer. Specifically for the \Sketch~task which has differing low-level features compared to ImageNet, the first convolution layer is consistently considered as task-specific. We see this is the case across varying values of $\lambda$, which aligns with the intuition that initial layers of ResNet should be retrained when transferring to a domain with different low-level features. A detailed figure of the task-specific layers for the \Sketch~task can be found in appendix \ref{fig:progress}. 

\paragraph{Effect of choice of pre-trained model}
To analyze the effect of using different pre-trained models, we replaced the base ImageNet model with a Places-365 model and applied \DWS{} on the datasets listed in \cref{tab:base}. We notice changes both in task-specific layers and in the performance. The number of task-specific layers increases for every task, in particular the average percentage of task-specific layers grows from $25.91\%$ to  $36.60\%$. We hypothesize that the Places-365 pre-training may not be well suited for object classification, so more layers need to be tuned. Supporting this, we also see a drop $2.77\%$ in average accuracy across the datasets (see \ref{tab:params-places} for details). These observations are consistent with the findings in \cite{mallya2018piggyback}.  

\begin{table}[h]
\centering
\caption{\textbf{Accuracy  of various methods across architectures and datasets}.  Classification accuracy for various methods is shown across  different datasets and architectures. The ViT-S/16 model has the highest accuracy, across datasets. \DWS{} is able to match the fine-tuning performance for ViT-S/16 and is about 1-2\% away for \DN{}.}
\begin{tabular}{lccccc}
\toprule
 & \Flowers & \Wiki & \Sketch  & \Cars & \CUB\\
\midrule
\textbf{\DN} & \\
 Fine-Tuning & 95.6 & 77.0  & 81.1 & 89.5	 & 82.6	 \\
Piggyback & 94.7 & 70.4 & 79.7 & 89.1 & 80.5 \\ 
 \DWS{}  &  95.8 & 73.6 & 80.2	 &  88.0 & 80.9\\

\midrule 
\textbf{ViT-S/16}
\\
Fine-Tuning            &99.3&82.6&81.9&89.2&88.9    \\
\DWS{}                    & 99.1&82.3&82.2&88.7&88.4    \\
\bottomrule
\end{tabular}
\label{tab:arch-perf}
\end{table}

\paragraph{Effect of Architecture Choice}
To demonstrate that \DWS{} is agnostic to architecture, we evaluate it on \DN{} \cite{densnet}. We report the performance of our method compared to fine-tuning and \PB~in \cref{tab:arch-perf}~and parameters in \cref{tab:arch-param}. The number of task-specific layers are high in \DN ~compared to \RN{50}. We conjecture that because of the extra skip connections, changing a single layer has more impact on the output compared to the ResNet model.

\begin{figure*}[h!]
    \centering
    \includegraphics[width=.93\linewidth]{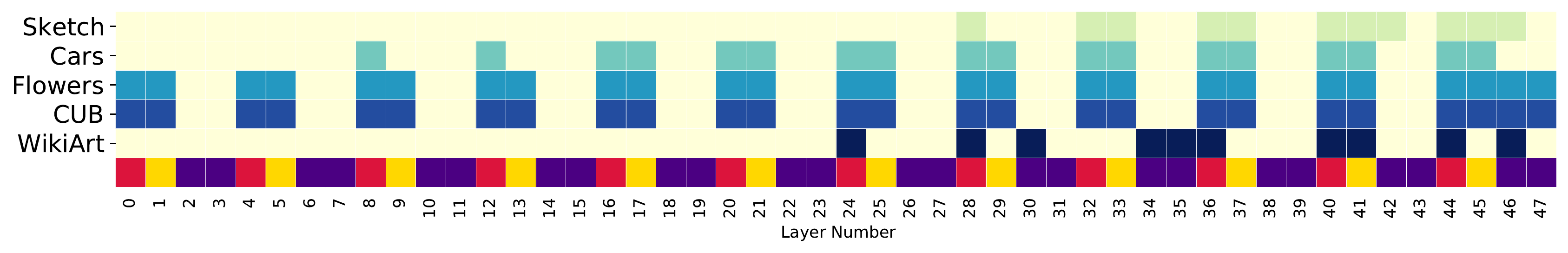}
    \caption{{\textbf{Task-specific layers for different datasets for the ViT model.} The figure shows the task-specific layers that are active for different datasets. Each rows shows the different layers that are present in a ViT model. Layer 0 is closest to the input. Layers shown in yellow are shared with the ImageNet pre-trained model. The last row shows the type of layer and is denoted in color. Here crimson denotes the Query-Key-Value in the attention layer, gold denotes the projection layer and purple denotes the MLP layer. Unlike CNNs, we see that the fine-tuning strategy is very different. Instead of freezing blocks, we need to freeze the MLP layers.  }}
    \label{fig:sharing_vit}
\end{figure*}

\paragraph{Transformers} We show results of our method on the transformer architecture. Here, we use the ViT-S/16 model \cite{steiner2021train} (see \ref{app:incMTL} for the training details). In \cref{tab:arch-perf} we report the performance of our method and parameters in \cref{tab:arch-param}. For the transformer architecture, the performance is better than CNNs as expected. We also notice that fewer parameters are made task-specific compared to CNNs. Although \DWS{} modifies more task specific layers, the adapted layers have fewer parameters.
We show the layers that are task-specific in \cref{fig:sharing_vit}. From this figure we see that the layers that are adapted to be task-specific for transformers follow a very different pattern from those of CNNs. While in the latter, lower layers tend to be task agnostic, and final layers task-specific, this is not the case for transformers. Here, layers throughout the whole network tend to be adapted to the task. Moreover, attention and projection layers tend to be adapted, whereas MLP layers are fixed. This shows that \DWS{} can adapt in nontrivial ways to different architectures without any hand-crafted prior.



\paragraph{Visual Decathlon}
\cref{tab:decathalon} shows that for $\lambda=0.25$, our method achieves the second highest S-score, without any dataset-specific hyper-parameter tuning. For this $\lambda$, we use half the number of parameters as Spot-Tune while performing better in 6/10 datasets and also have a higher mean score. All variants of our method outperform \RA{}, \DA{}, \PB{}. Moving from $\lambda=0.25$ to $\lambda=1.0$, we further reduce the number of task-specific layers by half while increasing the mean error only by $1\%$.  At $\lambda=1$, we we outperform \PB{} while using  lesser total number of parameters and storage space of the models, even considering  that \PB{} uses Boolean parameters.

\begin{figure}[h!]
    \centering
    \includegraphics[width=.95\linewidth]{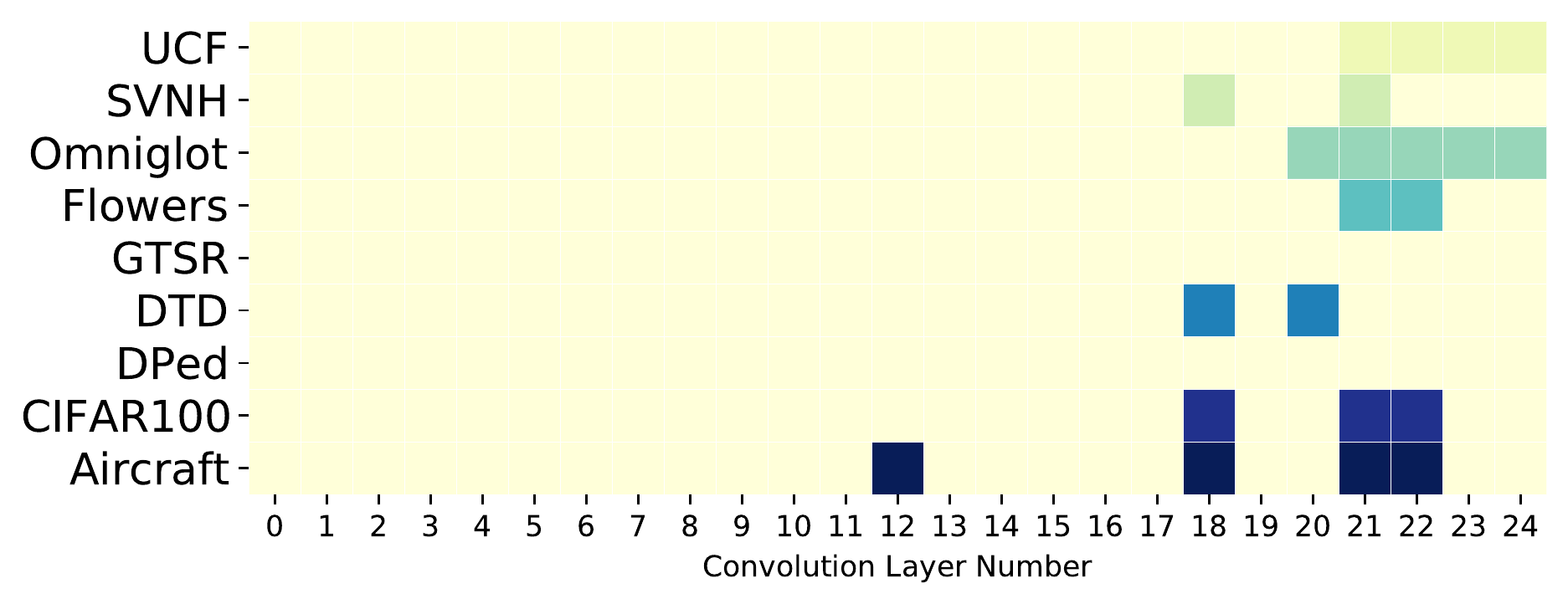}
    \caption{{\textbf{Task-specific layers for Visual Decathalon.} Each row shows the task-specific layers for different datasets ($\lambda=1.0$). For two datasets: DPed and GTSR, no task-specific parameters are needed. The performance improves solely due to updating the batch norm parameters. }}
    \label{fig:sharing_dec}
\end{figure}

To analyze what layers are being used we plot the active layers at the highest compression $(\lambda=1)$ in \cref{fig:sharing_dec}. For all datasets, the number of task-specific layers is small, with Omniglot requiring the most layers, while DPed and GTSR require the least. In fact, for the latter datasets, no task-specific layers are required outside of updating the batch norm layers (which leads to a significant boost in performance compared to fixed feature extraction). Again, we note that \DWS{} can easily find complex nonstandard sharing schemes for each dataset, which would otherwise have required an expensive combinatorial search. 
\begin{table*}[h]
\centering
\footnotesize
\caption{\textbf{Accuracy of various methods on the  Visual Decathlon Challenge datasets.} Accuracy for each dataset, the mean accuracy across all datasets and the S-Score \cite{rebuffi2017learning} is shown. \DWS{} has the second best S-Score at almost half the parameters of the best method. Our method can tradeoff accuracy vs additional parameters. We report the total number of parameters and  in  parenthesis the data-type normalized parameter count. }
\begin{tabular}{p{0.15\textwidth}|c|ccccccccc|c|c}
\toprule
Method & Params & Airc. & C100  & DPed & DTD & GTSR & Flwr. & Oglt. & SVHN & UCF & Mean & S-Score\\
\midrule
Fixed Feature  \cite{rebuffi2018efficient} &1$\times$ &  23.3 & 63.1 & 80.3 & 45.4 & 68.2 & 73.7 & 58.5 & 43.5 & 26.8 & 54.3 & 544 \\ 
FineTuning \cite{rebuffi2018efficient} & 10$\times$ &60.3  & 82.1 & 92.8 &55.5 & 97.5 & 81.4 & 87.7 & 96.6 & 51.2 & 76.5  & 2500	 \\
\midrule 
\RA \cite{rebuffi2017learning} & 2$\times$ & 56.7 & 81.2. & 93.9 & 50.9 & 97.1 & 66.2 & 89.6 & 96.1 & 47.5 & 73.9 & 2118 \\
DAM \cite{rosenfeld2018incremental} & 2.17$\times$ & 64.1 & 80.1 & 91.3 & 56.5 & 98.5 & 86.1 & \textbf{89.7} & 96.8 & 49.4 & 77.0& 2851 \\
PA \cite{rebuffi2018efficient} & 2$\times$& 64.2 & 81.9 & 94.7 & 58.8 & 99.4 & 84.7 & 89.2 & 96.5 & 50.9 & 78.1& 3412 \\
\PB \cite{mallya2018piggyback} & 10$\times$(3.25$\times$) & 65.3 & 79.9 & 97.0 & 57.5 & 97.3 & 79.1 & 87.6 & \textbf{97.2} & 47.5  &  76.6 & 2838\\
WTPB \cite{mancini2018adding} & 10$\times$(3.25$\times$)  &  52.8 & \textbf{82.0} & 96.2 & 58.7 & 99.2 & \textbf{88.2} & 89.2 & 96.8 & 48.6 & 77.2 & 3497 \\
BA$^2$ \cite {Berriel_2019_ICCV}  &6.13$\times$ (2.28$\times$) & 49.9 & 78.1 & 95.5 & 55.1 & 99.4 & 86.1 & 88.7 & 96.9& 50.2 & 75.7 & 3199\\
Spot-tune \cite{guo2019spottune} &  11$\times$ & 63.9 & 80.5 & 96.5 & 57.1 & \textbf{99.5 }& 85.2 & 88.8 & 96.7 & \textbf{52.3} & 78.1& 3612 \\
\midrule 
\DWS{}  ($\lambda$=0.25) &5.24$\times$&\textbf{66.58}&81.76&\textbf{97.07}&\textbf{58.83}&99.07&86.99&88.79&95.72&51.92&\textbf{78.70}&3533\\
\DWS{}  ($\lambda$=0.50) & 3.88$\times$&62.05&81.74&97.13&57.02&98.40&85.80&88.96&95.62&49.06&77.61&3180 \\

\DWS{}  ($\lambda$=0.75) & 3.43$\times$&62.62&81.07&95.77&57.34&98.61&85.67&89.00&95.65&49.56&77.56&3096\\

\DWS{}  ($\lambda$=1.0) & 3.13$\times$&63.43&81.04&96.99&58.19&98.38&84.08&89.16&94.99&51.10&77.77&3088\\

\bottomrule
\end{tabular}

\label{tab:decathalon}
\end{table*}

\subsection{Joint Multi-Task Learning}
\label{sec:mtl}
\begin{table*}[!htp]
\centering
\small
\caption{\textbf{\DWS{} vs AdaShare}. 
The accuracy of methods on the DomainNet dataset in both joint and incremental MTL settings is shown. All results are obtained with \RN{34} unless stated otherwise. Bold numbers represent the higher accuracy between \DWS{} and AdaShare. Numbers with underline denote the best performing method in each setting. \DWS{} outperforms AdaShare in both settings. The Params column measures the total parameters for supporting all tasks in comparison with the single base model.} 
\begin{tabular}{l|l|l|c|c|c|c|c|c|c}
\toprule
MTL Setting & Method            & Params & Real  & Painting & Quickdraw & Clipart & Infograph & Sketch & Mean  \\ \hline
\multirow{3}{*}{Joint } & Fine-tuning                           & 1$\times$     & 75.01 & 66.13    & 54.72     & 75.00      & 36.35     & 65.55  & 62.12 \\
& AdaShare                         & 1$\times$     & 76.90  & 67.90     & 61.17     & 75.88   & 31.52     & 63.96  & 62.88 \\
& \DWS{}  & 1.46$\times$  & \underline{\textbf{78.91}} & \underline{\textbf{67.91}}    & \underline{\textbf{70.18}}     & \underline{\textbf{76.98}}   & \underline{\textbf{39.30}}   & \underline{\textbf{67.81}}  & \underline{\textbf{66.84}} \\ 
\midrule
\multirow{3}{*}{Incremental } &{Fine-tuning} & 6$\times$     & \underline{81.51} & \underline{69.90}     & \underline{73.17}     & 74.08   & \underline{40.38}     & \underline{67.39}  & \underline{67.73} \\  
& AdaShare                     & 5.73$\times$    & 79.39 & 65.74    & 68.15     & 74.45   & 34.11     & 64.15  & 64.33 \\ 
& AdaShare \RN{50} &  4.99$\times$    & 78.71 &64.01    & 67.00     & 73.07   &    31.19  & 63.40 &  62.90\\
& \DWS{}  & 4.90$\times$  & \textbf{80.28} & \textbf{67.28}    & \textbf{71.79}    & \underline{\textbf{74.85}}   & \textbf{38.21}     & \textbf{66.66}  & \textbf{66.51} \\
\bottomrule
\end{tabular}


\label{tab:ada}
\end{table*}
\paragraph{Setting} 
We compare \DWS{} with AdaShare \cite{sun2019adashare} in the joint MTL setting, where multiple tasks are learned together with (selectively) shared backbones and independent task heads.
We follow the same setting as AdaShare, \ie, datasets and network for a fair comparison. 
Specifically, we compare performance on the DomainNet datasets~\cite{peng2019moment} with \RN{34}. This dataset contains the same labels across 6 domains and is an excellent candidate for MTL learning as there are opportunities for sharing as well as task competition.

To analyze the difference between \DWS{} and AdaShare, we further compare them in the incremental MTL setting, where each task is learned independently ($K=1$). We also include full fine-tuning as a baseline. Detailed training settings can be found in the \ref{app:jointMTL}.

\paragraph{Select-or-Skip vs Add-or-Not}
As shown in \Cref{tab:ada}, \DWS{} outperforms AdaShare in both the incremental and joint multi-task settings across all six domains. This may partly be due to AdaShare's select-or-skip strategy, which effectively reduces the number of residual blocks of the network. 
\DWS{}, on the other hand, replaces layers with task-specific versions without altering the model capacity, which results in performance closer to full fine-tuning.
To verify whether a larger network will improve AdaShare's performance, we perform the same incremental experiment with \RN{50}. 
Surprisingly, AdaShare with \RN{50} has a slightly lower performance than its \RN{34} version, suggesting that capacity might not be a limiting issue for the method. 


\paragraph{Incremental Training vs Joint Training} 
As shown in Table~\ref{tab:ada}, full fine-tuning for each task often yields the best performance and significantly outperforms the joint fine-tuning version.
This suggests that \emph{task competition} exists among the domains and only one domain (ClipArt)  benefits from the joint training. 
When Adashare is trained with the incremental setting, we also see an increase in performance compared to the joint fine-tuning version.
However, both methods come at the cost of weights sharing among tasks and a linear increase of the total model size.
For \DWS{}, we see the performance is relatively stable when switching from joint training to incremental training and is close to the performance of fine-tuning.



\paragraph{Parameter and Training Efficiency}
In the incremental setting, \DWS{} is 18\% more parameter efficient compared to Adashare while performs 2.18\% better on average. 
In this setting, AdaShare modifies the weights of existing layers and \emph{no} layers are shared across tasks. Parameter savings here come from skipped blocks. 
Conversely, in the joint MTL setting, AdaShare is more parameter efficient, as no new parameters are introduced. However, this leads to performance degradation.
\DWS{} performs uniformly better while introducing 0.46$\times$ more task-specific parameters. 

As for training efficiency, AdaShare learns the select-or-skip policy first, which optimizes both weights and policy scores alternatively. After the policy learning phase, multiple  architectures are sampled and retrained to get the best performance. 
This two-phase learning process increases the training cost significantly in comparison with \DWS{}, which has very little overhead compared to standard fine-tuning.



\section{Limitations}
A limitation of \DWS{} is that tasks do not share task-specific layers with each other, \ie, new tasks either learn their own task-specific parameters or share with the pre-trained model. The joint training approach we propose mitigates this issue by learning parameters common to all tasks. However, tasks added incrementally still cannot share parameters with others. Retraining the network on all tasks becomes infeasible as the number of tasks grow. We leave this aspect of parameter sharing as future work. 

\section{Conclusion}
We have presented \DWSF{}, a simple method to adapt a base model to a new task by modifying a small task-specific subset of layers. We show that we are able to learn which layers to share differentiably using a straight-through estimator with gating over task-specific weight deltas. Our experimental results show that, \DWS{} retains high accuracy on target tasks using task-specific parameters. \DWS{} is agnostic to the particular architecture used, as seen in our results with \RN{50}, \RN{34}, \DN{} and ViT models. We are able to discover standard and unique fine-tuning schemes. Furthermore, in the MTL setting we are able to avoid task competition  by using task-specific weights.

\clearpage
{\small
\bibliographystyle{ieee_fullname}
\bibliography{egbib}
}

\newpage

\appendix

In this supplementary material, we present additional details that we have referred to in the main paper.

\section{Datasets}
\label{app:datasets}
\paragraph{ImageNet to Sketch Dataset}
Originally introduced in \cite{mallya2018piggyback} and  referred to as  Imagenet to Sketch in \cite{Berriel_2019_ICCV}, our first benchmark consists of 5 datasets namely: \CUB{}\cite{birds}, \Cars{}\cite{cars}, \Wiki{}\cite{saleh2015large}, \Flowers{}\cite{flowers} and \Sketch{}\cite{eitz2012hdhso}. Following \cite{mallya2018packnet}, \Cars{} and \CUB{} are the cropped datasets, while the rest of the datasets are as is. 
For the \Flowers{} dataset, we combine both the train and validation split as the training data. 
We resize all images to 256 and use a random resized crop of size 224 followed by a random horizontal flip. 
This augmentation is applied at training time to all the datasets except \Cars{} and \CUB{}. 
For these two datasets, we use only the horizontal flip as the data augmentation during training as in \cite{mallya2018packnet,mallya2018piggyback,guo2019spottune}. 

The \Sketch{} dataset is licensed under a Creative Commons Attribution 4.0 International License. The CUBs, WikiArt, and Cars datasets use are restricted to non-commercial research and educational purposes.

\paragraph{Visual Decathlon Challenge}
The Visual Decathlon Challenge~\cite{rebuffi2017learning} aims at evaluating visual recognition algorithms on images from multiple visual domains. The challenge consists of  10 datasets: 
ImageNet~\cite{Russakovsky2015ImageNetLS}, Aircraft~\cite{aircrafts},  CIFAR-100~\cite{krizhevsky2009learning},  Describable textures~\cite{dtd}, Daimler pedestrian classification~\cite{munder2006experimental},  German traffic signs~\cite{stallkamp2012man},  UCF-101 Dynamic Images~\cite{soomro2012ucf101},  SVHN~\cite{svhn},  Omniglot~\cite{lake2015human}, and Oxford Flowers~\cite{flowers}. 
The images of the Visual Decathlon datasets are resized isotropically to have a shorter side of 72 pixels, to alleviate the computational burden for evaluation. Each  dataset has a different augmentation for training. We follow the  training protocol as used in \cite{rebuffi2017learning, rebuffi2018efficient}. 
Visual Decathlon does not explicitly provide a usage license.

\paragraph{DomainNet}
DomainNet~\cite{peng2019moment} is a benchmark for multi-source domain adaptation in object recognition. 
It contains 0.6M images across six domains (Clipart, Infograph, Painting, Quickdraw, Real, Sketch).
All domains include 345 categories (classes) of objects. 
Each domain is considered as a task and we use the official train/test splits in our experiments. We also use the same augmentations as used in \cite{sun2019adashare}. 
DomainNet is distributed under fair usage as provided for in section 107 of the US Copyright Law.



\section{Detailed Experimental Settings}
\subsection{Incremental MTL}
\label{app:incMTL}
As mentioned in the main paper,  we describe detailed settings for all the experiments in Sec 4.1.
\paragraph{\DN{} Training}
We sweep through learning rate of $\{0.005,0.01\}$ and use a similar setup to our \RN{50} model, \ie SGD optimizer with no weight decay. We train for 30 epochs and use a cosine annealing learning rate scheduler. We used the same augmentation as that of \RN{50} model as stated earlier. To summarize, the learning rate and the network architecture are the only things that change compared to the \RN{50} model training.

\paragraph{ViT-S/16 Training}
We train the ViT-S/16 model with a learning rate of 0.00005, momentum of 0.9, and $\lambda=1$. 
We train for 30 epochs with batch size of 32 and cosine annealing learning rate. Data augmentation consists of random resized crop and horizontal flip. 

\subsection{Joint MTL}
\label{app:jointMTL}
In this section, we describe detailed settings for the DomainNet experiments in Sec 4.2.
For each of the methods, we provide details regarding both joint and incremental MTL settings.

\paragraph{Fine-tuning}
For the joint MTL setting, fine-tuning is done with the adjustment of batch size and learning rate. The shared backbone is trained with a batch size of 72 with 12 images from each task, learning rate 0.005, momentum 0.9, and no weight decay. We train for 60 epochs where an epoch consists of sampling every image across all 6 datasets. 
We find that balanced sampling over all tasks for each mini-batch is necessary for stable training as in \cite{sun2019adashare}.

For the incremental MTL setting, we train 30 epochs with batch size 32, learning rate 0.005, momentum 0.9, and weight decay 0.0001.

\paragraph{AdaShare}
The learning process of AdaShare consists of two phases: the policy learning phase and retraining phase.
During the policy learning phase, both weights and policies are optimized alternatively for 20K iterations.
After the policy learning phase, different architectures are sampled with different seeds and the best results are reported.
For the joint MTL setting, we follow the same setting as the original implementation~\cite{sun2019adashare}. 
Note that the original implementation samples 8 architectures from the learned policy and retrains them and report the best result, here for fair comparison with other methods, only 1 sampling is performed.
For the incremental setting, we use the same setting as the joint-MTL, except the dataset contains only 1 domain. 

\paragraph{\DWS}
For all experiments we use learning rate 0.005, momentum 0.9, no weight decay, and $\lambda=1$. 
Data augmentation consists of a random resized crop and horizontal flip. 
For both the joint and incremental setting we train with a batch size of 32 and 30 epochs. The difference being for the joint variant we report in the main paper we start with the jointly trained backbone then train with the incremental variant of TAPS as opposed to an generic model for incremental training. When we use the joint version in which the backbone is trained simultaneously with the weight deltas, we train with batch size 72 and sample 12 images per task for each batch for 60 epochs.



\section{Additional Results}
\subsection{Imagenet to Sketch Benchmark}

\paragraph{Amount of Additional Parameters}
\begin{table}[h]
\centering
\small
\begin{tabular}{c c c c c}
\toprule
 \Flowers & \Wiki & \Sketch  & \Cars  & \CUB\\
\midrule
\textbf{\% Addl. Params.}\\
65.50 & 52.82 & 75.87 & 41.85 & 70.61\\
\bottomrule
\end{tabular}
\caption{ \textbf{Additional parameters for \RN{50} model on ImageNet-to-Sketch benchmark}. Amount of Additional parameters (percentage) for \DWS{}{} are shown for the  different datasets.}
\label{tab:base-param}
\end{table}
\cref{tab:base-param} shows the percentage of additional parameters needed for each dataset of this benchmark. In total we use about $4.12\times$ the number of parameters, with \Cars{} using the least parameters and \Sketch{} using the most. 

\begin{figure*}[h!]
    \centering
    \includegraphics[width=.99\linewidth]{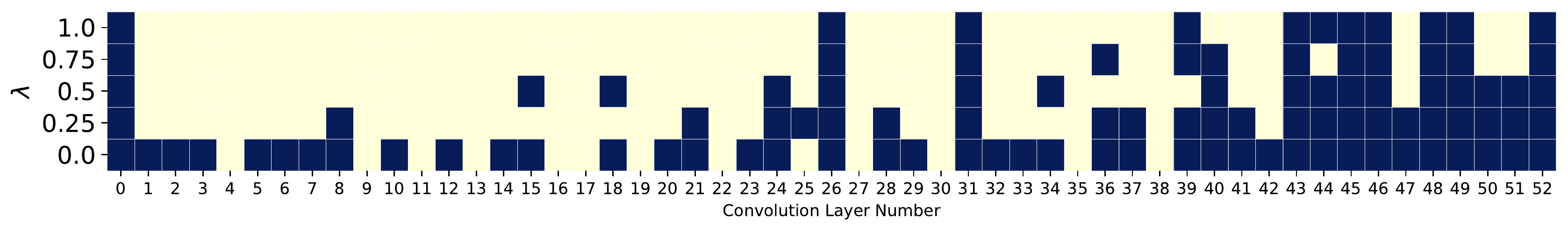}
    \caption{\textbf{When do we use task specific weights?.} The plot shows the different task specific layers for the a model trained on the \Sketch{} dataset. Each row represents the value of $\lambda$ for which the network uses task specific weights. Blue boxes indicate task specific layers while yellow boxes indicate base model layers. From this figure we can see that the lowest layers are active for any value of $\lambda$. Most other lower layers are turned off when $\lambda>0$}
    \label{fig:progress}
\end{figure*}

\paragraph{Layers active for the \Sketch{} dataset for various $\lambda$}
We show in \cref{fig:progress}, the layers that are active for various values of $\lambda$ for the \Sketch{} dataset. From this figure, we see that the first layer is task specific across different values of $\lambda$. This shows that the first layer is crucial for fine-tuning. This intuitively makes sense as the \Sketch{} dataset has very different low level statistics compared to ImageNet. 

\paragraph{Effect of Pretraining on number of Task specific Parameters}
\begin{table*}[h!]
\centering
\small
\begin{tabular}{llllll}
\toprule
                       & \Flowers & \Wiki & \Sketch & \Cars & \CUB \\ \hline
                        Resnet - 50 & \\
\% Addl. Parameters & \\
 Imagenet &  65.50 & 52.82 & 75.87 & 41.85 & 70.61\\
 
 Places  &  75.08 & 68.72 & 74.59 & 75.02 & 74.72\\
\hline
\multicolumn{4}{l}{\% Task Specific Layers} & \\
 Imagenet  & 22.64 & 20.75& 43.40& 14.47 & 28.30 \\
 Places  & 35.85 & 28.30 & 50.94 & 33.96 & 33.96 \\
\hline
\end{tabular}
\caption{\textbf{Effect of Pretrained Model on Task specific parameters.} Amount of parameters that are additionally needed for each task. The comparison between Imagenet pretrained model and Places pretrained model is shown. Across the board we see that the Places pretrained model uses more parameters as well as layers. }
\label{tab:params-places}
\end{table*}
Show in \cref{tab:params-places} is the number of task specific layers and parameters that are needed for each dataset. From this table, we see that for all datasets the number of layers that are tasks specific are more for Places pretraining compared to an Imagenet pretraining. One could attribute this to the fact that the  Places model specializes in scenes whereas the Imagenet model in objects. Hence, the Imagenet model is "closer" to the tasks as opposed to the Places model. 

\begin{figure}[h!]
    \centering
    \includegraphics[height=.99\textheight]{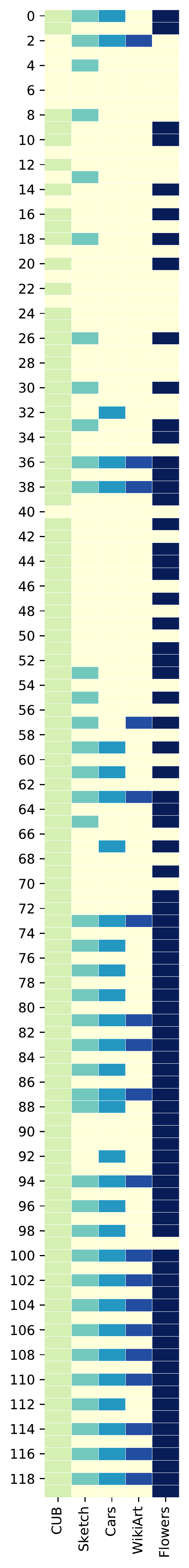}
    \caption{{\textbf{Shared layers for different tasks.} The figure shows the task specific layers that are active for different datasets using a \DN{} model. We observe that compared to the \RN{50} model, many more layers are active for the same dataset.}}
    \label{fig:sharing_dense} 
\end{figure}

\paragraph{Layers active for \DN{}}
\cref{fig:sharing_dense} shows the layers that are active for the \DN{} model. From this figure we see that the number of layers that are task specific are much greater compared to the \RN{50} model. Our hypothesis is that since \DN{} has more skip connections, changing a layer has an effect on more layers as compared to \RN{50}.

\subsection{Visual Decathalon}
\label{app:VisualDec}
\begin{table*}[h]
\centering
\small
\begin{tabular}{p{0.2\textwidth}cccccccccccc}
\toprule
Method  & Airc. & C100  & DPed & DTD & GTSR & Flwr. & Oglt. & SVHN & UCF & Mean. & S-Score\\
\midrule
\DWS  ($\lambda$=1.0) &  63.43 & 81.04 & 96.99 & 58.19 & 98.38 & 84.08 & 89.16 & 94.99 & 51.10 & 77.77 & 3088 \\
\% Addl. Parameters  & 32.95&30.43&0.13&20.38&0.13&20.33&47.45&20.41&40.53\\
\% Task specifc layer   & 16.00 & 12.00 & 0.00 & 8.00 & 0.00 & 8.00 & 20.00 & 8.00 & 16.00 \\
\midrule 
\DWS ( ($\lambda$=0.25) &66.58&81.76&97.07&58.83&99.07&86.99&88.79&95.72&51.92& 78.703& 3532\\
\% Addl. Parameters & 80.92&60.73&0.13&53.28&20.38&40.53&66.38&40.69&60.72\\
\% Task specific layer  &  44.00&24.00&0.00&24.00&8.00&16.00&28.00&16.00&24.00 \\
\bottomrule
\end{tabular}
\caption{ \textbf{Visual Decathalon additional parameter count.} Shown in this table is the percentage of task specific layers as well as parameters needed for each task in addition to the base model. We show this for $\lambda=1.0$ and $\lambda=0.25$} 
\label{tab:dec-param-count}
\end{table*}
In \cref{tab:dec-param-count}, we show the percentage of task specific layers as well as parameters for $\lambda=1.0$ and $\lambda=0.25$. From this table we see that as $\lambda$ increases, the percentage of task specific layers and parameters decreases. For the datasets where there are no task specific layers, the increase in parameters is  due to the storage of batch norm parameters. 

\begin{figure}[h!]
    \centering
    \includegraphics[width=.99\linewidth]{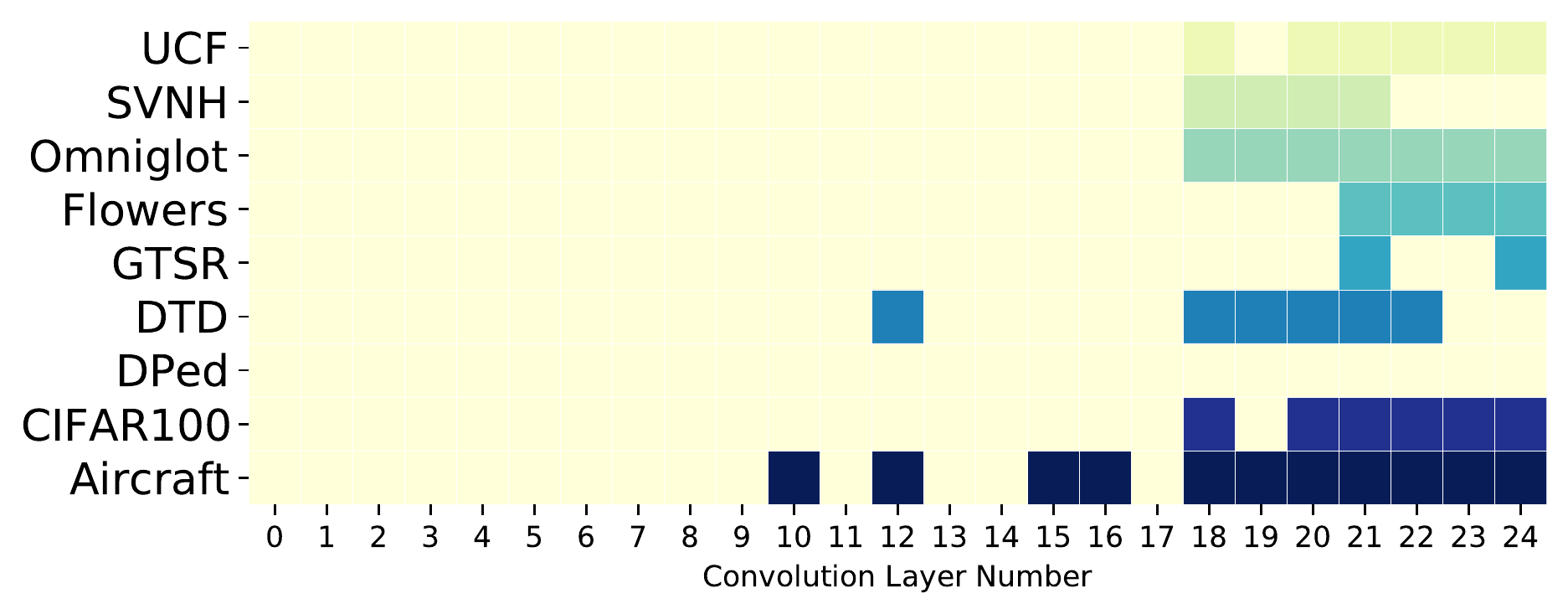}
    \caption{{\textbf{Task specific layers for different tasks.} The figure shows the task specific layers that are adapted for different datasets in the Visual Decathlon challenge for $\lambda=0.25$}}
    \label{fig:sharing25} 
\end{figure}

\begin{figure}[h!]
    \centering
    \includegraphics[width=.99\linewidth]{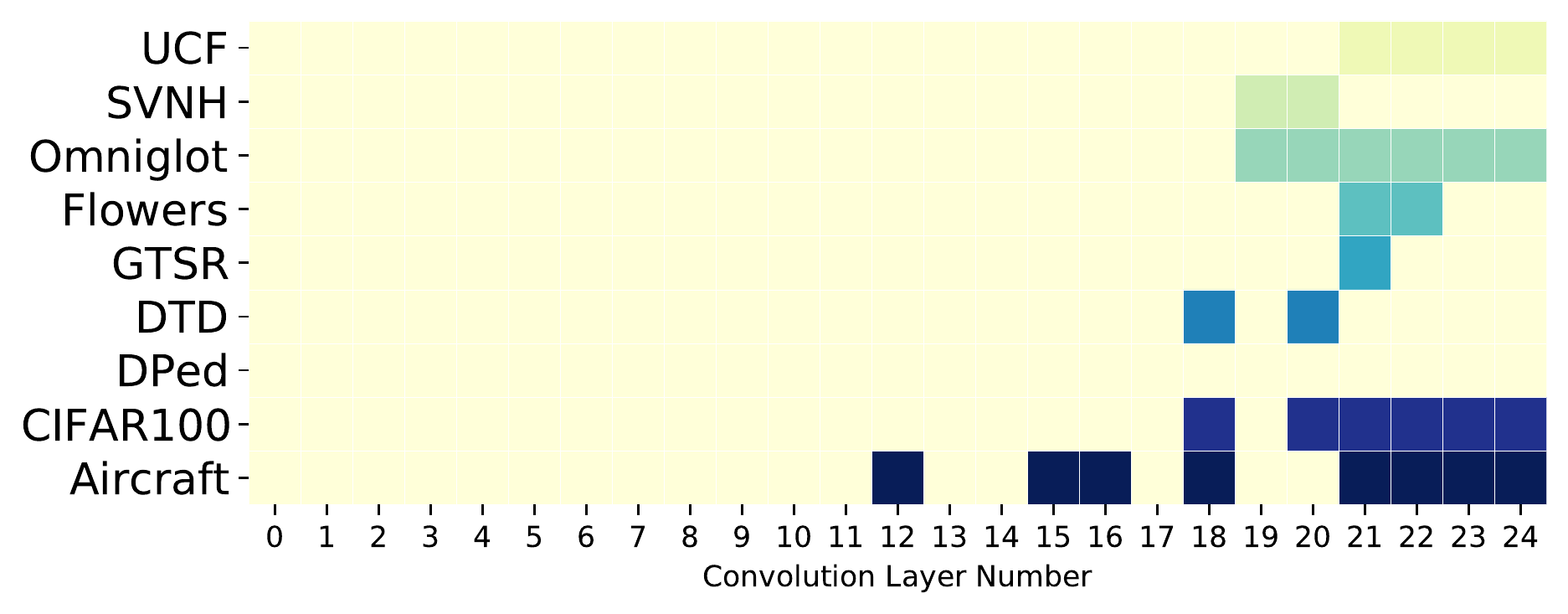}
     \caption{{\textbf{Task specific layers for different tasks.} The figure shows the task specific layers that are adapted for different datasets in the Visual Decathlon challenge for $\lambda=0.50$}}
   \label{fig:sharing50} 
\end{figure}

\begin{figure}[h!]
    \centering
    \includegraphics[width=.99\linewidth]{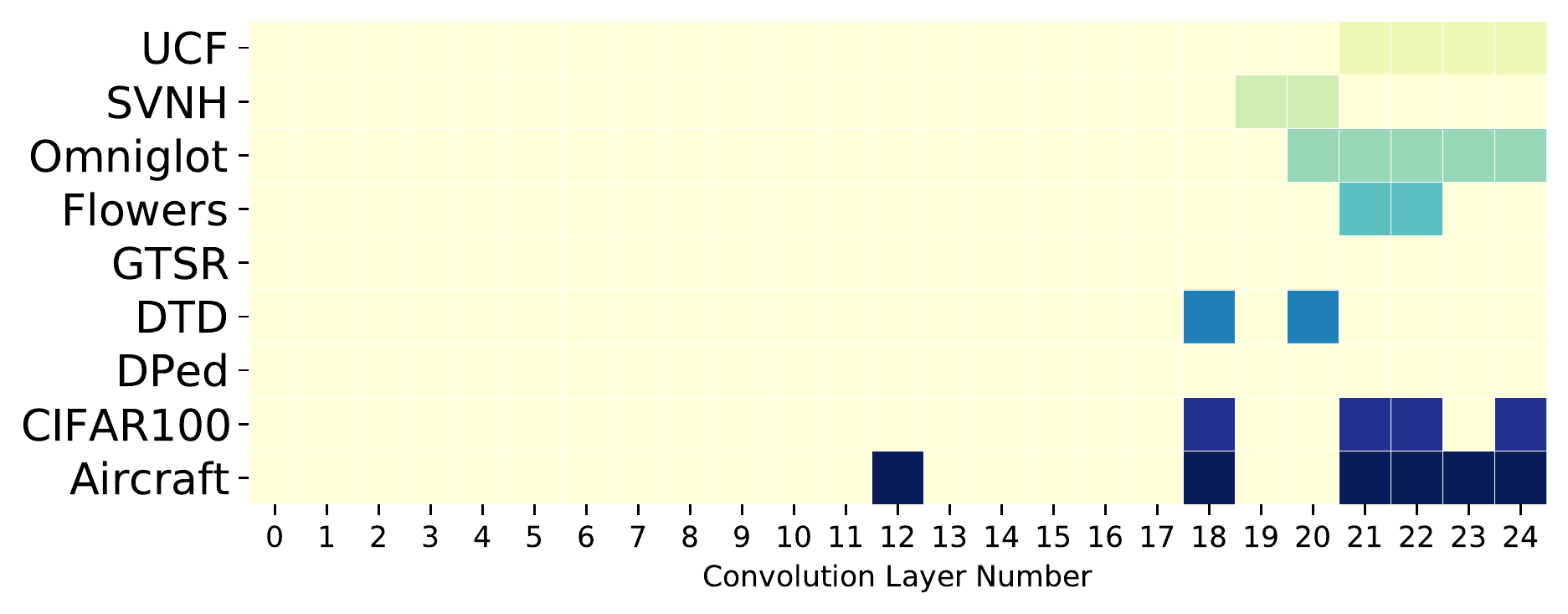}
  \caption{{\textbf{Task specific layers for different tasks.} The figure shows the task specific layers that are adapted for different datasets in the Visual Decathlon challenge for $\lambda=0.75$}}
   \label{fig:sharing75} 
\end{figure}

\begin{figure}[h!]
    \centering
    \includegraphics[width=.99\linewidth]{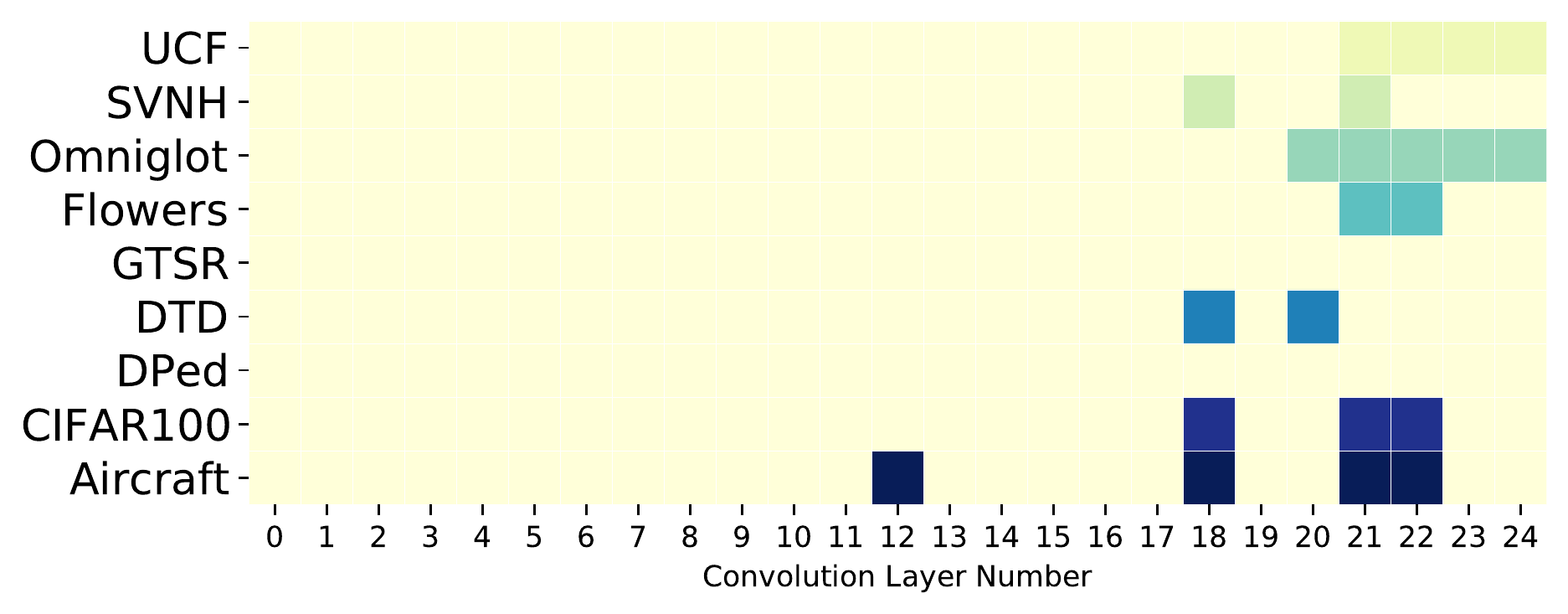}
  \caption{{\textbf{Task specific layers for different tasks.} The figure shows the task specific layers that are adapted for different datasets in the Visual Decathlon challenge for $\lambda=1.0$}}
   \label{fig:sharing100} 
\end{figure}

We also show the layers where task specific adaptation is needed for different values of $\lambda$ for the Visual Decathlon dataset. This can be seen in \cref{fig:sharing25} $(\lambda=0.25)$, \cref{fig:sharing50} $(\lambda=0.5)$, \cref{fig:sharing75} $(\lambda=0.75)$ and \cref{fig:sharing100} $(\lambda=1.0)$. For all the different values of $\lambda$, we see that almost all of the layers below layer 9 are not adapted.  For the Aircraft dataset, layer 12 is consistently adapted. Similarly for the DTD dataset we consistently observe that no layers  are adapted. This shows that some adaptations are independent of $lambda$ and hence critical for the task. 

\section{Memory Efficient Joint Variant}
See table \ref{tab:jointVar} for comparison between the memory efficient variant and standard \DWS{} on the DomainNet benchmark for joint multi-task learning.
\label{sec:JointVariant}
\begin{table*}[h!]
\centering
\begin{tabular}{l|l|l|c|c|c|c|c|c}
\toprule
~Method            & Params & Real  & Painting & Quickdraw & Clipart & Infograph & Sketch & Mean  \\ \hline
\multirow{3}{*} ~\DWS{}~(Standard)                      & 1.43$\times$     & 78.45 & \underline{\textbf{68.23}}    & \underline{\textbf{70.32}}     & \underline{\textbf{77.00}}      & \underline{\textbf{39.35}}     & 67.95  & \underline{\textbf{66.88}} \\
~\DWS{}~(Mem. Efficient)  & 1.46$\times$  & \underline{\textbf{78.91}} & 67.91    & 70.18     & 76.98   & 39.30   & \underline{\textbf{67.81}}  & 66.84 \\
\bottomrule

\end{tabular}
\caption{\textbf{Comparison between the memory efficient and standard version of \DWS{} on DomainNet in the joint MTL setting.} The memory efficient version performs comparably in accuracy and parameter cost while only needing constant memory during training.}
\label{tab:jointVar}
\end{table*}

\section{Batch Norm and Manual Freezing}
In Tab. \ref{tab:rebuttalbn} we report the results of only changing batch norm parameters and manually freezing layers to match the parameter cost of \DWS{}. We find that adaptively modifying layers outperforms manual freezing. 
\section{PyTorch Implementation}
We show the code snippet of a PyTorch implementation of \DWS~in Algorithm~\ref{code:indicator} and Algorithm~\ref{code:taps}. 
The AdaptiveConv conv layers can be used to replace the normal Conv2d layers in an existing model. This shows how easily existing architectures can be used for training \DWS{}.

\onecolumn{ 
\definecolor{codeblue}{rgb}{0.25,0.5,0.5}
\definecolor{codeblue2}{rgb}{0,0,1}
\lstset{
  backgroundcolor=\color{white},
  basicstyle=\fontsize{12pt}{12pt}\ttfamily\selectfont,
  columns=fullflexible,
  breaklines=true,
  captionpos=b,
  commentstyle=\fontsize{10pt}{10pt}\color{codeblue},
  keywordstyle=\fontsize{10pt}{10pt}\color{codeblue2},
}

\begin{algorithm}[h!]
\caption{\large Pytorch Code for Gating Function with Straight Through Estimator}
\begin{lstlisting}[language=Python]
class BinarizeIndictator(autograd.Function):
    @staticmethod
    def forward(ctx, indicator, threshold=0.1):
        out = (indicator >= threshold).float()
        return out
    @staticmethod
    def backward(ctx, g):
        # send the gradient g straight-through on the backward pass.
        return g, None

\end{lstlisting}
\label{code:indicator}
\end{algorithm}

\begin{algorithm}[h!]
\caption{\large Pytorch Code for Incremental Version of Task-Adaptive Convolutional Layers}
\begin{lstlisting}[language=Python]

class AdaptiveConv(nn.Conv2d):
    def __init__(self, *args, **kwargs):
        super().__init__(initial_val, *args, **kwargs)
        weight_shape = self.weight.shape
        #Initialize residual weights and indicator scores.
        self.residual = torch.nn.Parameter(torch.zeros(weight_shape))
        self.indicator = torch.nn.Parameter(torch.ones([initial_val])
        #Freeze base network weights
        self.weight.requires_grad = False
        self.weight.grad = None


    def forward(self, x):
        I = BinarizeIndictator.apply(self.indicator)
        w = self.weight + I * self.residual
        x = F.conv2d(x, w)
        return x
        
\end{lstlisting}
\label{code:taps}
\end{algorithm}
}

\begin{table}[h!]
\centering
\begin{tabular}{l|l|ccccc}
\toprule
  & Param & \Flowers & \Wiki & \Sketch  & \Cars  & \CUB\\
\midrule
Feat. Extractor & 1$\times$ &  89.14 & 61.74	 & 65.90 & 55.52 & 63.46 \\
BN & 1.01$\times$ &  91.07 & 70.06 & 78.47 & 79.41 & 76.25 \\ 
Man. Freeze & 4.15$\times$ &  92.35 & 73.32 & 78.68 & 87.62 & 81.01 \\ 
TAPS & 4.12$\times$ & 96.68& 76.94&80.74&89.76&82.65\\
\bottomrule
\end{tabular}
\centering
\caption{\textbf{Performance of manually freezing layers and only adapting the batch norm parameters with a \RN{50}~ model on ImageNet-to-Sketch benchmark.} We observe that adapatively selecting which layers to modify with \DWS{} outperforms manually freezing an equivalent number of parameters starting from the last layers.}
\label{tab:rebuttalbn}
\end{table}

\twocolumn

\end{document}